\documentclass{article}

\usepackage[final]{nips_2017}

\usepackage[utf8]{inputenc} 
\usepackage[T1]{fontenc}    
\usepackage{hyperref}       
\usepackage{url}            
\usepackage{booktabs}       
\usepackage{amsfonts}       
\usepackage{nicefrac}       
\usepackage{microtype}      
\usepackage{amsmath}
\usepackage{tabularx}
\usepackage{tabu}

\usepackage{mathtools}
\usepackage{nicefrac}

\title{Probability Series Expansion Classifier that is Interpretable by Design }

\author{
  Sapan Agarwal\\
  Sandia National Laboratories\\
  PO Box 969 MS 9158\\
  Livermore, CA 94551-0969 \\
  \texttt{sagarwa@sandia.gov} \\
  \And
  Corey M. Hudson \\
  Sandia National Laboratories \\
  PO Box 969 MS 9291 \\
  Livermore, CA 94551-0969\\
  \texttt{cmhudso@sandia.gov} \\
}

\begin{document}
\maketitle
\begin{abstract}
This work presents a new classifier that is specifically designed to be fully interpretable. This technique determines the probability of a class outcome, based directly on probability assignments measured from the training data. The accuracy of the predicted probability can be improved by measuring more probability estimates from the training data to create a series expansion that refines the predicted probability.   We use this work to classify four standard datasets and achieve accuracies comparable to that of Random Forests. Because this technique is interpretable by design, it is capable of determining the combinations of features that contribute to a particular classification probability for individual cases as well as the weightings of each of combination of features.
\end{abstract}
\section{Introduction}
When exploring the decisions a machine-learning classifier makes, it is often challenging to understand the individual choices that contribute to the decision, for all but the most trivial classifiers [1-3]. This is increasingly problematic as machine-learning classifiers handle more critical [4] and socially important systems [5], and as governments [6] and citizens demand more oversight [7]. The problem emerges because most classifiers are trained through the solution of a complex optimization problem that is designed to fit a large number of model parameters. In order to avoid over-fitting the data, the data is regularized and learning rules are extracted, thereby reducing the overall information content in hopes of achieving relevant, rather than exact models [8]. This process inherently obscures the properties of the training data that leads to the classification, resulting in the need for heuristic methods to try to interpret the classifier. Simple decision tree based classifiers have a straight-forward interpretation, but in order to get high classification accuracies and reduce over-fitting, the results of many trees need to be combined, thus obscuring any attempts to interpret the model.

To overcome these limitations, we propose a new classifier that estimates the probability of a classification outcome based directly on probabilities measured from the training data. We will show that this classifier has an accuracy performance that is comparable to random forests. Additionally, this classifier allows us to determine the combinations of features that contributed to the particular classification as well as their weights.

\section{Series Classifier Model}
The goal of the new classifier is to predict an outcome $Y = Y'$, given measured categorical features $x_1 = x_1',\: x_2 = x_2',\: x_3 = x_3'$,… In particular, we want to estimate: 
\begin{equation} \label{eq:eq1}
P( Y = Y'|x_1 = x_1',x_2 = x_2',x_3 = x_3', \ldots )
\end{equation}
using only probabilities measured from the training data. We seek to measure the probability that $Y = Y'$ given different combinations of features such as: $P(Y = Y')$, $P(Y = Y'|x_1 = x_1')$, $P(Y = Y'|x_3 = x_3')$, $P(Y = Y'|x_1 = x_1',\:x_2 = x_2')$, … The key challenge is deciding how to combine these measured probabilities into a singular estimate. Unfortunately, there are no good first principles methods for combining these probabilities. Trying to use Bayesian methods to put limits on Eq. \eqref{eq:eq1} will often give limits of 0 and 1, giving no useful information. The simplest option is to average probabilities:
\begin{equation} \label{eq:eq2}
P( Y = Y'|x_1 = x_1',x_2 = x_2',x_3 = x_3', \ldots ) \approx \left. { \sum \nolimits_i P(Y = Y'|x_i = x_i')} \middle/ { \sum \nolimits_i 1} \right.
\end{equation}
However, this will not account for a number of effects as described in the following sections. To account for those effects, we will hierarchically weight the averages of the contributing probabilities. First, we define the following notation:
\begin{subequations} \label{eq:eq3}
\noindent\begin{minipage}{.4\linewidth}
\begin{align}
	P &= P(Y = Y')\label{eq:eq3a} \\ 
	P_i &= P(Y = Y'|x_i= x_i')\label{eq:eq3b}
\end{align}
\end{minipage}%
\begin{minipage}{.6\linewidth}
\begin{align}
	P_{ij} &= P( Y = Y'|x_i= x_i',\: x_j= x_j')\label{eq:eq3c} \\
	P_{ijk} &= P(Y = Y'|x_i= x_i',\: x_j = x_j',\: x_k = x_k') \label{eq:eq3d}
\end{align}
\end{minipage}
\end{subequations}
and so on. Furthermore, we define $P'$ as the estimate of $P$ after accounting for $P_i'$, $P_i'$ is the estimate of $P_i$ after accounting for $P_{ij}'$ and so on.

\subsection{Create a Series Expansion of Probabilities}
To account for multiple levels of measured probabilities, the probability estimate can hierarchically be refined as follows:

\begin{subequations} \label{eq:eq4}
\noindent\begin{minipage}{.33\linewidth}
\begin{equation}\label{eq:eq4a}
		P' = \frac{ \sum \nolimits_i {P_i'}} { \sum \nolimits_i 1}
\end{equation}
\end{minipage}%
\begin{minipage}{.33\linewidth}
\begin{equation}\label{eq:eq4b}
  		P_i' = \frac{ \sum \nolimits_{j \ne i} P_{ij}'} { \sum \nolimits_{j \ne i} 1}
\end{equation}
\end{minipage}
\begin{minipage}{.33\linewidth}
\begin{equation} \label{eq:eq4c}
		P_{ij}' = \frac{ \sum \nolimits_{k \ne j \ne i} P_{ijk}'} { \sum \nolimits_{k \ne j \ne i} 1}  
\end{equation}
\end{minipage}
\end{subequations}

The overall probability estimate is evaluated, starting at the lowest level ($P(Y = Y')$ given the largest number of features) for which there is a measured probability and working upwards until the estimate $P'$ is computed. The sums are simply taken over probabilities that have been measured in the training data. If an estimated probability such as $P_{ij}'$, is not known, the directly measured value ${P_{ij}}$ should be used where available.

If all possible combinations of features are kept, the size of the model would grow exponentially as the number of levels in the model increases. To limit the model size, we selectively choose the probabilities to keep track of, based on which combinations of features minimize Gini Impurity (or entropy). In the tests performed later, we kept all possible probabilities, $P_i$ and $P_{ij}$ for the first two levels and then kept only the top six $P_{ijk}$ for each $P_{ij}$ and the top six $P_{ijkl}$ for each $P_{ijk}$ used.
Missing features can easily be handled by leaving out probabilities corresponding to the missing features.
\subsection{Weight P=0 and P=1 More Strongly}
If any feature defines $P=0$ or $P=1$, then the overall probability is 0 or 1, regardless of other features. Consequently, when averaging probabilities in Eq. \eqref{eq:eq4}, the closer a probability is to 1 or 0, the more it should be weighted. Thus, we introduce the following scaling weight $S_i$:
\begin{equation} \label{eq:eq5}
	S_i = \left. {1} \middle/ [{P_i' \times ( 1 - P_i' ) }] \right.
\end{equation}
The probability computation Eq. \eqref{eq:eq4a} is updated as follows:
\begin{equation} \label{eq:eq6}
P' = { \sum_i {P_i' \times S_i}} \bigg / { \sum \nolimits_i {S_i}} 
\end{equation}
A logarithmic scaling that increases more slowly as $P$ approaches 0 or 1 was included as a heuristic in earlier versions of the algorithm, but was found to lower classifier performance.
\subsection{Account for Noise and Low Data Counts}
Each probability estimate, $P_i$, $P_{ij}$, etc. is backed by some number of training examples, $n_{train}$. If this number is too low, noise can result in a poor probability estimate. It is also possible to get $P=0$ or $P=1$ for some combinations of features, which could cause the scaling factor Eq. \eqref{eq:eq5} to blow up. To account for this we need to make three corrections: 1) Eq. \eqref{eq:eq5} needs to be updated to account for the number of training examples 2) Each probability estimate should be weighted by the amount of data backing it and 3) If there is not enough data backing a probability estimate its parent estimate should be used (i.e. use $P_i$ instead of $P_{ij}$)

First, we need to limit the $P_i'$ used in Eq. \eqref{eq:eq5} based on the number of training samples backing $P_i$:

\begin{subequations} \label{eq:eq7}
\begin{equation} \label{eq:eq7a}
S_i = \left. {1} \middle/ [P_i^{''} \times ( 1 - P_i^{''} )] \right.
\end{equation}
\begin{equation} \label{eq:eq7b}
P_i^{''} = 
\begin{dcases} 
    0.5 &  n_{train,i} \le n_{err}\\
    {n_{err}} / {N_{train,i}} & P_i' < {n_{err}} / {N_{train,i}} \text{ and } n_{train,i} > n_{err}\\
    P_i' & {n_{err}} /{N_{train,i}}  \le P_i' \le 1 - {n_{err}} /{N_{train,i}}\text{ and } n_{train,i} > n_{err}\\
	1 - {n_{err}} /{N_{train,i}} & 	P_i' > 1 - {n_{err}} /{N_{train,i}} \text{ and }n_{train,i} > n_{err}
\end{dcases}
\end{equation}
\end{subequations}
Here $n_{err}$ is a is a meta-parameter that represents average number of examples that might be wrong due to noise.  Next, we need to add a weight to the averages in Eq. \eqref{eq:eq3} based on the amount of data used:
\begin{subequations} \label{eq:eq8}
\begin{equation}
P' = \frac{ \sum_i P_i' \times S_i \times N_i}  {\sum_i S_i \times N_i}
\end{equation}
\begin{equation}
N_i = \min\left( \sqrt {n_{train,i}} ,\sqrt{n_{max}}  \right)
\end{equation}
\end{subequations}
Here $n_{max}$ is a meta-parameter that represents the number of training examples needed to overcome noise. Finally, if none of the probabilities are backed by significant data, the parent probability should be used:
\begin{equation} \label{eq:eq9}
 P' = \frac{ \max \limits_i  (N_i) }  {N}  \times
      \frac{\sum_i P_i' \times S_i \times N_i } {\sum_i S_i \times N_i} + 
	  \left( 1 - \frac{ \max \limits_i (N_i)} {N} \right) \times P
\end{equation}

\subsection{Account for the Usefulness of a Feature Combination}
If a particular combination of features significantly changes the probability estimate (i.e. $P_{ij}' >> P_i$) it can contain new information and should therefore be weighted more heavily. Consequently, we can optionally add a usefulness weight, $U_i$:
\begin{equation} \label{eq:eq10}
U_i = (P_i' - P) \:/ \:(P_i' + P)
\end{equation}
Using this weight was found to improve accuracy for some datasets, but not all.
\begin{table}[t]
  \caption{The new series classifier achieves accuracies comparable to Random Forests}
  \label{table1}
  \centering
  \begin{tabular}{lll}
    \toprule
    Dataset     & Series Classifier     & Random Forests \\
    \midrule
    1984 Congressional Voting Records [11,12] 	& 96.8\%  & 96.8\%  \\
    E. coli Promotor Gene Sequences [12,13]     & 95.3\%  & 94.3\%  \\
    SPECT Heart Data [12,14]     				& 85.0\%  & 85.0\%  \\
    Lymphography [12,15]						& 86.4\%  & 87.2\%  \\
    \bottomrule
  \end{tabular}
\end{table}

\subsection{Overall Model}
The overall model accounting for all the above factors is given by:
\begin{subequations} \label{eq:eq11}
\begin{align}
P' &= \frac{\max_i ( N_i )} {N} \times 
	 \frac{\sum_i P_i' \times S_i \times N_i \times U_i} {\sum_i S_i \times N_i \times U_i} +
	 \left(1 - \frac{\max_i ( N_i )} {N}  \right) \times P\\
P_i' &= \frac{\max_j ( N_{ij} )} {N_i} \times 
	 \frac{\sum_{j \ne i} P_{ij}' \times S_{ij} \times N_{ij} \times U_{ij}} 
	 {\sum_{j \ne i} S_{ij} \times N_{ij} \times U_{ij}} +
	 \left(1 - \frac{\max_j ( N_{ij} )} {N_i}  \right) \times P_i\\
P_{ij}' &= \frac{\max_k ( N_{ijk} )} {N_{ij}} \times 
	 \frac{\sum_{k \ne j \ne i} P_{ijk}' \times S_{ijk} \times N_{ijk} \times U_{ijk}} 
	 {\sum_{k \ne j \ne i} S_{ijk} \times N_{ijk} \times U_{ijk}} +
	 \left(1 - \frac{\max_k ( N_{ijk} )} {N_{ij}}  \right) \times P_{ij}
\end{align}
\end{subequations}
\section{Classification Results}
To benchmark the performance of this new classifier, we compared classification accuracy against scikit-learn’s [9] Random Forests implementation [10] on four different datasets (summarized in Table \ref{table1}). As seen in the table the new classifier achieves accuracies comparable to Random Forests. The hyper parameters ($n_{err}$, $n_{max}$, tree depth, and use/don’t use usefulness model) were optimized using 4-fold cross-validation and the best accuracies are reported. 

\section{Interpretability Results}
Because this classifier assesses combinations of probabilities, it can directly calculate the probability contributed to the overall classification estimate and use this to understand which features contributed most highly to its estimate. To demonstrate this, we analyze an example in the congressional voting records dataset that was misclassified in Table \ref{table2}. The classifier misclassified with 70\% probability that this Member of Congress would be a Republican when they are a Democrat. Table \ref{table2} shows the top six feature combinations that contributed to the classification and the corresponding weight assigned to each feature. The particular votes that the features correspond to are given in the Supplementary Information.  This shows that in the training data certain feature combinations indicated with 100\% probability that the Member of Congress was a Republican and other feature combinations indicated that the Member of Congress was a Democrat with 100\% probability. To understand the impact of each individual feature, given the feature combinations, the probability estimates, $P_i'$ and the corresponding weights are shown in Table \ref{table3}. These features indicated that the Member of Congress was a Republican and were weighted more highly resulting in the classification of this member as a Republican.
\begin{table}[t]
  \caption{Analysis of a misclassified instance by feature combination}
  \label{table2}
  \centering
  \begin{tabularx}{\textwidth}{X p{1.3cm} p{1.3cm} p{1.4cm} p{1.2cm} p{1.4cm}}
    \toprule
    Features 	& Counts      	& Counts  	& Probability  	& Weight 	& Cumulative \\
         		& Republican  	& Democrat 	& Republican 	&  			&  Weight\\

    \midrule
    $x_3=1$, $x_9=1$, $x_{15}=0$ 		& 16 	& 0 	& 100\%  	& 12.5\% 	& 12.5\%\\
    $x_9=1$, $x_{15}=0$ 			& 16  	& 0 	& 100\%  	& 7.0\% 	& 19.4\%\\
	$x_3=1$, $x_8=0$, $x_9=1$, $x_{14}=0$	& 47	& 0		& 100\%		& 4.8\%		& 24.2\% \\
	$x_3=1$, $x_7=0$, $x_9=1$, $x_{14}=0$	& 47	& 0		& 100\%		& 4.7\%		& 28.9\% \\
	$x_2=1$, $x_{10}=1$, $x_{11}=0$		& 0		& 58	& 0\%		& 4.0\%		& 32.9\% \\
	$x_1=1$, $x_2=1$, $x_{10}=1$		& 0		& 38	& 0\%		& 3.1\%		& 36.0\% \\

    \bottomrule
  \end{tabularx}
\end{table}

\begin{table}[t]
  \caption{Analysis of a misclassified instance by individual feature}
  \label{table3}
  \centering
  \begin{tabular}{lccccc}
    \toprule
	Feature \# & Probability Republican & Weight &Feature \# & Probability Republican & Weight \\
    \midrule
	$x_9=1$ 	& 95\% 	& 17\% 	& $x_8=0$	& 86\%	& 4\% \\
	$x_3=1$	& 94\%	& 16\%	& $x_7=0$	& 83\%	& 4\% \\
	$x_{15}=0$	& 94\%	& 14\%	& $x_6=0$	& 29\%	& 3\% \\
	$x_2=1$	& 14\%	& 8\%	& $x_0=0$	& 78\%	& 3\% \\
	$x_{10}=1$	& 23\%	& 8\%	& $x_{13}=1$	& 55\%	& 2\% \\
	$x_{11}=0$	& 21\%	& 6\%	& $x_4=1$	& 69\%	& 2\% \\
	$x_{14}=0$	& 89\%	& 5\%	& $x_{12}=1$	& 52\%	& 2\% \\
	$x_1=1$	& 53\%	& 4\%	& $x_5=1$	& 56\%	& 2\% \\

    \bottomrule
  \end{tabular}
\end{table}
\section{Conclusion}
Specifically designing a classifier for interpretability naturally allows for more highly interpretable results. We show that is possible to design a classifier based on averaging measured probabilities that gives classifications which are as accurate as those in Random Forests. Using this classifier, we can easily determine which features and combinations of features contribute to a classification and audit and assess the limitations and strength of the model, identify potential problems in the training data and explain outlying instances. 

\subsubsection*{Acknowledgments}

Sandia National Laboratories is a multimission laboratory managed and operated by National Technology and Engineering Solutions of Sandia, LLC., a wholly owned subsidiary of Honeywell International, Inc., for the U.S. Department of Energy’s National Nuclear Security Administration under contract DE-NA0003525.

\section*{References}

[1]	L. Pulina and A. Tacchella. An Abstraction-Refinement Approach to Verification of Artificial Neural Networks. In Proc. 22nd Int. Conf. on Computer Aided
Verification (CAV), pages 243–257, 2010.

[2]	L. Pulina and A. Tacchella. Challenging SMT Solvers to Verify Neural Networks. AI Communications, 25(2): 117-135, 2012.

[3]	G. Katz et al., "Reluplex: An Efficient SMT Solver for Verifying Deep Neural Networks," Presented at CAV 2017. arXiv:1702.01135, 2017.

[4]	Z. Kurd, et al., “Developing Artifical Neural Networks for Safety Critical Systems,” Neural Computing and Applications, vol. 16, no. 1, pp. 11-19, 2007.

[5]	M. Skirpan and T. Yeh. “Designing a Moral Compass for the Future of Computer Vision using Speculative Analysis” Published in Computer Vision and Pattern Recognition Workshops, 2017.

[6]	B. Shneiderman. “The Dangers of Faulty, Biased, or Malicious Algorithms Requires Independent Oversight.” PNAS, vol. 113, no. 48, pp. 13538-13540, 2016.

[7]	J. Larson, et al., “Machine Bias: Investigating Algorithmic Injustice,” Series Published on ProPublica, https://www.propublica.org/series/machine-bias/, Accessed 2017.

[8]	D. MacKay. “Information-based Objective Functions for Active Data Selection,” Neural Computation, vol. 4, no. 4, pp. 590-604, 1992.

[9]	F. Pedregosa et al., "Scikit-learn: Machine learning in Python," Journal of Machine Learning Research, vol. 12, no. Oct, pp. 2825-2830, 2011.

[10]	L. Breiman, “Random Forests”, Machine Learning, vol. 45, no. 1, pp. 5-32, 2001.

[11]	Congressional Quarterly Almanac, “98th Congress, 2nd session 1984”, Congressional Quarterly Inc, vol. XL, 1985. 

[12]	M. Lichman, “UCI Machine Learning Repository”, http://archive.ics.uci.edu/ml, Accessed 2017.

[13]	G. Towell, et al., “Refinement of Approximate Domain Theories by Knowledge-Based Artificial Neural Networks,” Proceedings of the Eighth National Conference on Artificial Intelligence (AAAI-90), 1990.

[14]	K. Cios, et al,. “Hybrid inductive machine learning: An overview of CLIP algorithms,” New Learning Paradigms in Soft Computing, pp. 276-322, 2002

[15]	R. Michalski, et al., “The multi-purpose incremental learning system AQ15 and its testing application to three medical domains.” Proceedings of the Eighth National Conference on Artificial Intelligence (AAAI-86), 1986. 

\newpage

\section*{Supplementary Information}
Each of the features listed in Tables \ref{table2} and \ref{table3} correspond to votes on the following bills.  A vote of yes corresponds to a feature value of 1 and a vote of no corresponds to a feature value of 0.

\begin{table}[t]
  \caption{Bill descriptions for each feature number}
  \label{tableSI}
  \centering
  \begin{tabular}{ll}
    \toprule
	Feature Number & House Bill Description\\
    \midrule
0	& Handicapped infants\\
1	& Water project cost sharing\\
2	& Adoption of the budget resolution\\
3	& Physician fee freeze\\
4	& El Salvador aid\\
5	& Religious groups in schools\\
6	& Anti-satellite test ban\\
7	& Aid to Nicaraguan contras\\
8	& Mx missile\\
9	& Immigration\\
10	& Synfuels corporation cutback\\
11	& Education spending\\
12	& Superfund right to sue\\
13	& Crime\\
14	& Duty free exports\\
15	& Export administration act South Africa\\
    \bottomrule
  \end{tabular}
\end{table}

\end{document}